\documentclass[10pt,twocolumn,letterpaper]{article}

\usepackage[pagenumbers]{cvpr} 

\usepackage[utf8]{inputenc}         %
\usepackage[T1]{fontenc}            %
\usepackage{url}                    %
\usepackage{booktabs}               %
\usepackage{amsfonts}               %
\usepackage{nicefrac}               %
\usepackage{microtype}              %
\usepackage[dvipsnames]{xcolor}     %

\usepackage[utf8]{inputenc}         %
\usepackage[T1]{fontenc}            %
\usepackage{url}                    %
\usepackage{booktabs}               %
\usepackage{amsfonts}               %
\usepackage{nicefrac}               %
\usepackage{microtype}              %
\usepackage{xcolor}                 %

\usepackage[flushleft]{threeparttable}
\usepackage{float}
\usepackage{graphicx}
\usepackage{multirow}
\usepackage{xspace}
\usepackage{natbib}
\usepackage{enumitem}
\usepackage[font=small]{caption}
\usepackage{autobreak}
\usepackage{sidecap}
\usepackage{wrapfig}
\usepackage[toc, page, header]{appendix}
\usepackage{makecell}

\newcommand{\opt}{TRivia\xspace}

\newcommand{\redtext}[1]{\textcolor{red}{#1}}

\usepackage{tabularx}
\usepackage{soul, color, xcolor}
\usepackage{adjustbox}
\usepackage{colortbl}
\usepackage{CJKutf8}
\usepackage{tcolorbox}
\usepackage{hhline}
\usepackage[accsupp]{axessibility}  

\definecolor{cvprblue}{rgb}{0.21,0.49,0.74}
\usepackage[pagebackref,breaklinks,colorlinks,allcolors=cvprblue]{hyperref}

\def\paperID{} 
\def\confName{CVPR}
\def\confYear{2026}

\title{\opt: Self-supervised Fine-tuning of Vision-Language Models\\for Table Recognition}

\author{
Junyuan Zhang$^{1*}$\quad
Bin Wang$^{2*}$\quad
Qintong Zhang$^{3}$\quad
Fan Wu$^{2}$\quad
Zichen Wen$^{2,4}$\quad
Jialin Lu$^{1}$\quad\\
Junjie Shan$^{1}$\quad
Ziqi Zhao$^{1}$\quad
Shuya Yang$^{1}$\quad
Ziling Wang$^{1}$\quad
Ziyang Miao$^{2}$\quad
Huaping Zhong$^{5}$\quad\\
Yuhang Zang$^{2}$\quad
Xiaoyi Dong$^{2}$\quad
Ka-Ho Chow$^{1\ddag}$\quad
Conghui He$^{2\ddag}$\quad\\
$^1$The University of Hong Kong\quad
$^2$Shanghai AI Laboratory\quad\\
$^3$Peking University\quad
$^4$Shanghai Jiaotong University\quad
$^5$Sensetime\quad
}

\begin{document}
\maketitle
\renewcommand{\thefootnote}{}
\footnotetext{$^*$ These authors contributed equally to this work.}
\footnotetext{$^\ddag$ Corresponding Authors: Ka-Ho Chow, Conghui He.}
\begin{abstract}
Table recognition (TR) aims to transform table images into semi-structured representations such as HTML or Markdown.
As a core component of document parsing, TR has long relied on supervised learning, with recent efforts dominated by fine-tuning vision-language models (VLMs) using labeled data.
While VLMs have brought TR to the next level, pushing performance further demands large-scale labeled data that is costly to obtain.
Consequently, although proprietary models have continuously pushed the performance boundary, open-source models, often trained with limited resources and, in practice, the only viable option for many due to privacy regulations, still lag far behind.
To bridge this gap, we introduce \opt, a self-supervised fine-tuning method that enables pretrained VLMs to learn TR directly from unlabeled table images in the wild. 
Built upon Group Relative Policy Optimization, \opt automatically identifies unlabeled samples that most effectively facilitate learning and eliminates the need for human annotations through a question-answering-based reward mechanism. 
An attention-guided module generates diverse questions for each table image, and the ability to interpret the recognition results and answer them correctly provides feedback to optimize the TR model.
This closed-loop process allows the TR model to autonomously learn to recognize, structure, and reason over tables without labeled data. 
Leveraging this pipeline, we present \opt-3B, an open-sourced, compact, and state-of-the-art TR model that surpasses existing systems (e.g., Gemini 2.5 Pro, MinerU2.5) on three popular benchmarks.
Model and code are released at: \url{https://github.com/HKU-TASR/TRivia}
\end{abstract}

\begin{figure}[t]
    \centering
    \includegraphics[width=0.9\linewidth]{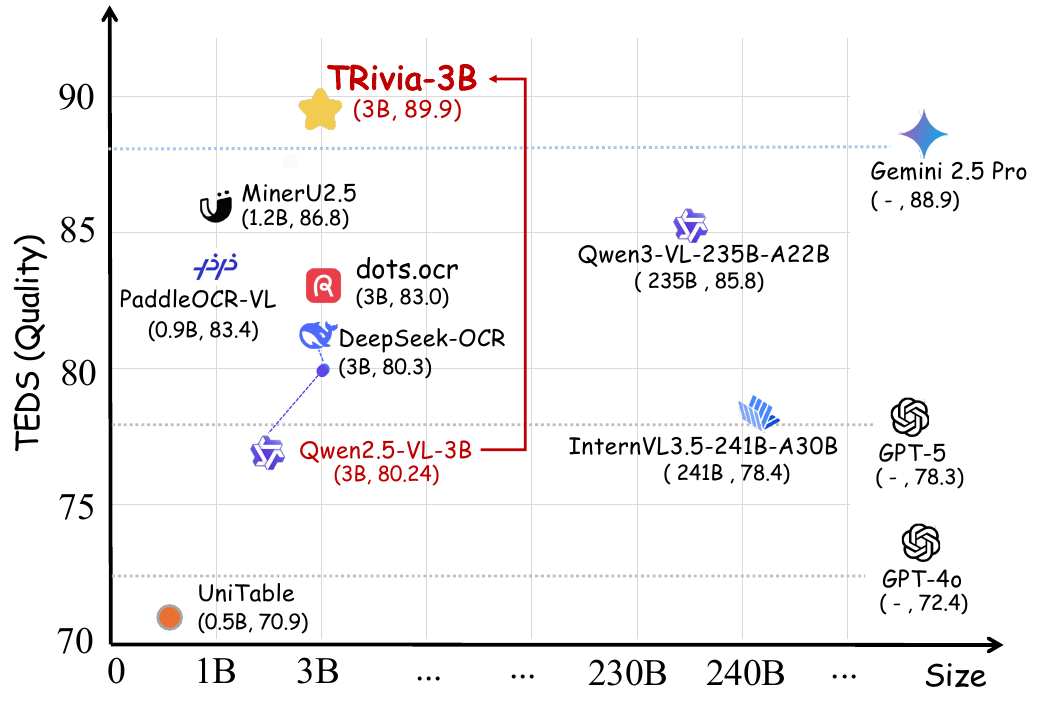}
    \vspace{-2mm}
    \caption{
    \opt-3B learns from unlabeled table images and achieves TR quality beyond the limit attainable by fine-tuning with labeled data.
    Unlike proprietary systems such as Gemini~2.5~Pro, it is open-sourced, compact, and can be deployed offline for privacy-sensitive document processing.
    }
    \vspace{-3mm}
    \label{fig:intro}
\end{figure}

\section{Introduction}

Document parsing plays a pivotal role in digitalization by converting scanned or photographed documents into machine-readable formats~\cite{zhang2024document,wang2024mineru} for downstream tasks such as retrieval-augmented generation~\cite{zhang2024ocr,hui2024uda}. 
Among document elements, tables remain a longstanding challenge.
They are both information-dense and structurally complex, requiring not only accurate text extraction, as in standard optical character recognition (OCR), but also precise reconstruction of their spatial and logical organization into semi-structured representations such as HTML or Markdown.

Recent advances in vision-language models (VLMs)~\cite{bai2025qwen2,comanici2025gemini,wang2025internvl3,hurst2024gpt,wen2025stop} have revolutionized table recognition (TR)~\cite{li2025monkeyocr,liu2025points,poznanski2025olmocr,chen2025logics,niu2025mineru2}. 
General-purpose VLMs can be adapted to TR either through prompt engineering or fine-tuning with image-annotation pairs. Proprietary systems such as Gemini~2.5~Pro, built with massive human and computational resources, already exhibit unprecedented TR capabilities as indicated in Figure~\ref{fig:intro}. 
Yet, they are accessible only through commercial APIs, raising privacy and compliance concerns in many practical scenarios involving sensitive documents.
In contrast, open-source VLMs offer greater flexibility and transparency but are constrained by the limited scale of existing TR datasets~\cite{huang2023improving,long2021parsing}, which are insufficient to reach state-of-the-art performance comparable to proprietary models.
For example, UniTable, trained solely on open-source data~\cite{huang2023improving,long2021parsing}, performs poorly across real-world benchmarks.
Even large-scale open-source efforts such as MinerU2.5, which combines millions of samples, costly human annotations, and distillation from proprietary models, still inherit the ceiling imposed by its teacher, Gemini~2.5~Pro.

Existing approaches to data acquisition for TR follow three paradigms. 
(1) Synthetic data, generated by rendering HTML tables, provides perfect annotations and scalability~\cite{li2025monkeyocr,feng2025dolphin,liu2025points} but lacks the visual diversity and domain alignment of real-world data~\cite{liu2025points}. 
(2) Real-world data better capture the complexity of practical documents, but are expensive and time-consuming to annotate~\cite{chen2025logics}. 
(3) Distilling pseudo-labels from proprietary models offers a potential compromise~\cite{poznanski2025olmocr}, yet this approach remains costly, inherently caps performance at that of the teacher model, and may violate service agreements.
Consequently, the curation of large-scale labeled datasets has become a major bottleneck, even though it is a critical factor for success.
This raises an important question: \emph{Can unlabeled table images from the wild be harnessed to improve TR performance?}

In this paper, we present \opt, a self-supervised fine-tuning framework that enables VLMs to learn table recognition directly from unlabeled table images, which can be collected at scale.
As not all samples contribute equally to learning, identifying those that are most informative and deriving verifiable supervisory signals without human intervention are key challenges.
\opt leverages Group Relative Policy Optimization (GRPO)~\cite{shao2024deepseekmath} with a decoupled reward function~\cite{xing2025caprl,qiao2024prism} to strengthen TR capabilities via unlabeled data.
Given a base VLM and a pool of unlabeled table images, it features a response-consistency sampling strategy that builds on GRPO’s principle of optimizing relative reward differences across multiple responses to select the most informative samples for learning. 
\opt then generates supervisory signals through a proxy task: table question-answering (QA). 
An attention-guided QA generation module produces diverse, verifiable questions about each table image, and the VLM is fine-tuned to recognize, structure, and reason over tables by producing recognition results that allow a language model to correctly answer these questions. 

The main contributions of this paper are as follows.
\begin{itemize}
	\item We investigate self-supervised fine-tuning of VLMs on unlabeled table images and propose \opt, a framework that harvests such data to push the frontier of TR.
	\item We introduce a response-consistency sampling strategy to automatically identify unlabeled samples that most effectively enhance VLM fine-tuning via GRPO.
	\item We design an attention-guided QA mechanism to ensure diverse, verifiable, and stable supervisory signals for optimization.
\end{itemize}
Based on the proposed framework, we present \opt-3B, an open-source, state-of-the-art TR model fine-tuned from Qwen2.5-VL-3B using unlabeled table images.
As shown in Figure~\ref{fig:intro}, \opt-3B surpasses both expert TR models (e.g., MinerU2.5 and PaddleOCR-VL) and general-purpose models (e.g., Gemini~2.5~Pro and GPT-5). 
We envision that \opt will open a new avenue for advancing TR beyond the limitations of labeled data via self-supervised fine-tuning.

\begin{figure*}[t]
    \centering
    \includegraphics[width=0.9\linewidth]{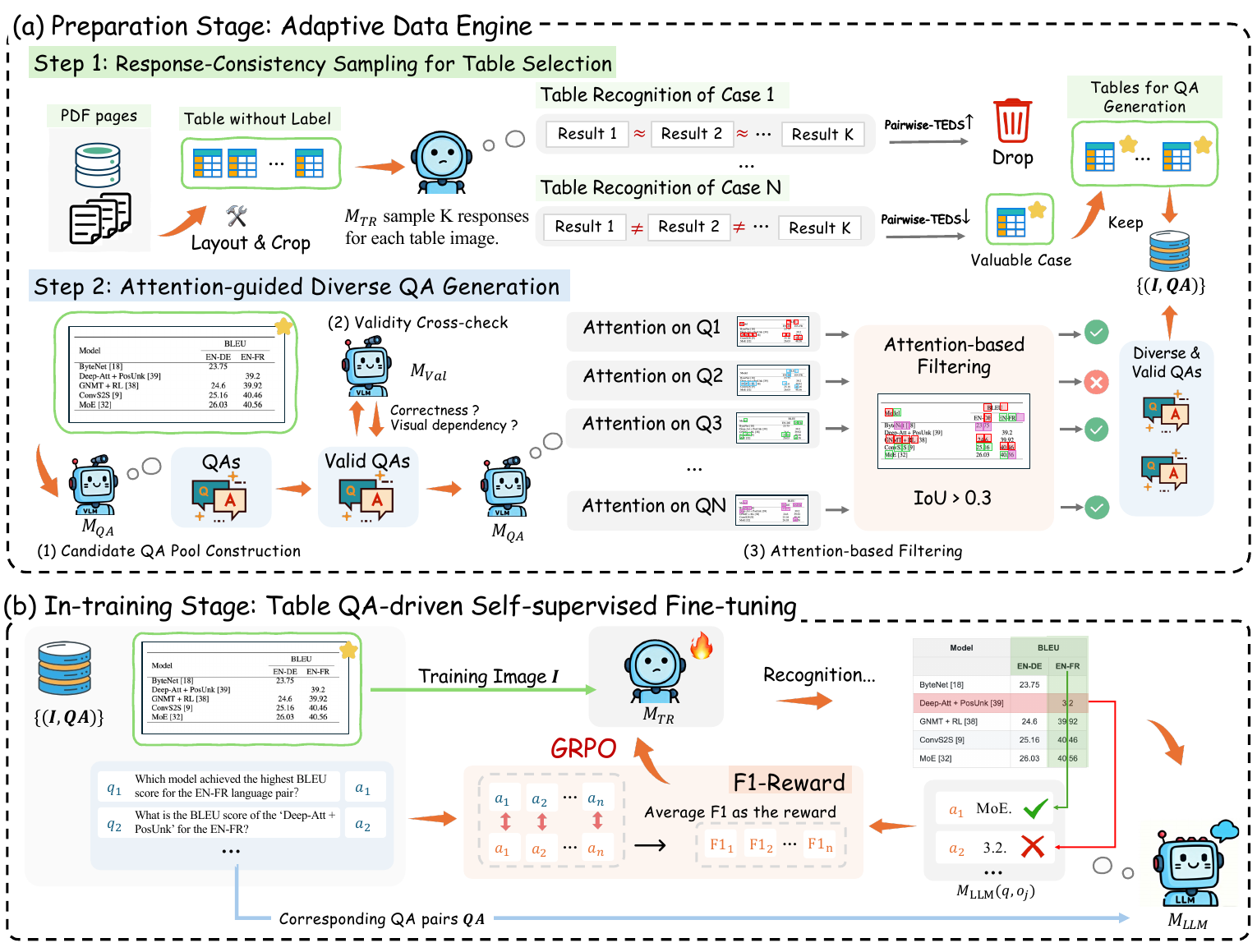}\vspace{-0.6em}
    \caption{
    \opt features (a) an adaptive dataset curation module to set the stage for (b) reinforcement learning to learn TR from unlabeled data.
    During dataset curation, \opt uses a response-consistency sampling strategy (Section~\ref{sec:sampling}) to identify informative samples and generate verifiable, diverse QA for each image through an attention-guided module (Section~\ref{sec:qa_generation}).
    Based on curated data, \opt fine-tunes the VLM to recognize, structure, and reason over tables through QA-based rewards (Section~\ref{sec:rl_training}).
    }
    \vspace{-6mm}
    \label{fig:pipeline}
\end{figure*}
\section{Related Work}
Early research on TR typically decomposes the task into two subtasks: table structure recognition and OCR. 
For structure recognition, bottom-up approaches~\cite{liu2022neural,raja2020table,xing2023lore} treat text or cell bounding boxes as graph nodes and predict their row, column, or cell associations. 
Split-and-merge methods~\cite{tensmeyer2019deep,zhang2022split,zhang2024semv2,lin2022tsrformer} first partition a table into a uniform grid and then merge adjacent cells to reconstruct its structure. 
While effective on clean layouts, these modular pipelines suffer from cumulative error propagation and rely heavily on explicit visual cues, making them brittle under real-world conditions such as distorted layouts, borderless designs, or complex merged cells~\cite{long2021parsing}.
These limitations have motivated the development of image-to-markup methods that reformulate TR as an end-to-end sequence-generation problem, directly converting table images into formats such as HTML~\cite{zhong2020image,nassar2022tableformer,huang2023improving,peng2024unitable}.
However, these models are constrained by their context window, input resolution, and training data quality, limiting their ability to handle large or structurally complex tables.
For instance, UniTable~\cite{peng2024unitable} processes images up to $448\times448$ with a 512-token output cap, resulting in notable performance drops on complex real-world tables.

Recent advances in VLMs have demonstrated remarkable generalization across OCR-related tasks~\cite{fu2024ocrbench,liu2024ocrbench,ouyang2025omnidocbench,yang2025cc,wen2026innovator,wen2025efficient}. 
By incorporating table recognition data during pretraining, general-purpose VLMs such as Gemini~2.5~Pro~\cite{comanici2025gemini} and the Qwen2.5-VL series~\cite{bai2025qwen2} can perform TR directly through natural-language instructions, eliminating the need for explicit OCR or cell-detection modules. 
Building on this trend, several task-specific expert VLMs~\cite{wei2024general,niu2025mineru2,feng2025dolphin,li2025monkeyocr,dotsocr,liu2025points} have been developed explicitly for TR, typically fine-tuned on large synthetic datasets augmented with a small number of real tables annotated either manually~\cite{chen2025logics,niu2025mineru2} or by proprietary systems~\cite{poznanski2025olmocr,niu2025mineru2}.
Nevertheless, even large-scale open-source efforts such as MinerU2.5~\cite{niu2025mineru2}, which combines millions of samples, costly human annotations, and distilled labels from Gemini~2.5~Pro, remain constrained by the performance ceiling of its teacher model, as shown in Figure~\ref{fig:intro}.
In contrast, our work departs from reliance on labeled or distilled data and uses table QA-based rewards as proxy supervision with GRPO to learn directly from unlabeled real-world table images.

\section{Methodology}
Figure~\ref{fig:pipeline} provides an overview of \opt, which consists of two stages: (i) a preparation stage that prepares supervisory signals from unlabeled table images, and (ii) an in-training stage that formulates TR learning as a semi-supervised problem. 
We first describe how \opt fine-tunes the VLM via GRPO using table QA as proxy rewards in Section~\ref{sec:rl_training}. 
We then introduce the adaptive data engine in Section~\ref{sec:data_engine}, which identifies informative samples and generates supervisory signals tailored to our fine-tuning strategy and the base VLM.

\subsection{Table QA-driven Self-supervised Fine-tuning}\label{sec:rl_training}
To push TR performance beyond the limits of supervised fine-tuning on labeled data, we adopt reinforcement learning (RL), which updates the model based on a reward function rather than explicit labels. 
As long as a reliable reward can be derived from unlabeled data, the model can still benefit from meaningful supervision and learn to claim the reward.

Table question-answering (QA) can serve as a proxy task to provide annotation-free reward signals because, as a downstream application of TR, the ability to correctly answer questions about a table implicitly reflects how well the recognized table preserves both textual and structural information~\cite{zhang2024ocr}. 
This is beneficial for two reasons.
First, generating valid QA pairs from a table image is significantly easier than predicting its full HTML markup: the model does not need to explicitly infer complex structures such as colspan and rowspan, but only to reason about the content of specific regions.
Second, the quality of QA pairs, including correctness and visual dependency, can also be cross-checked using additional models, which is not yet feasible when using HTML as labels.
These properties make table QA a reliable reward for self-supervised fine-tuning.

\noindent\textbf{Table QA-driven GRPO.} \opt adopts Group Relative Policy Optimization (GRPO) as the reinforcement learning framework to fine-tune the base VLM, such as the one optimized using existing labeled data, and to be further enhanced by learning from unlabeled data.
The overall GRPO training pipeline with QA-based rewards is illustrated in Figure~\ref{fig:pipeline}(b). 
Formally, each training sample is represented as $(\boldsymbol{I}, \boldsymbol{QA})$, where $\boldsymbol{I}$ denotes a table image and $\boldsymbol{QA}=\{(\boldsymbol{q}, \boldsymbol{a})\}$ represents its associated QA set. 
The process for generating QAs from $\boldsymbol{I}$ is detailed in Section~\ref{sec:data_engine}. 
During GRPO training, the fine-tuning-in-progress TR model $M_{\text{TR}}$ acts as the policy and produces a group of $R$ recognition responses ${\boldsymbol{o}_1, \boldsymbol{o}_2, \ldots, \boldsymbol{o}_{R}}$ for each image $\boldsymbol{I}$. 
Each recognition output $\boldsymbol{o}_j$ is then paired with table question $\boldsymbol{q}$'s and passed to a large language model (LLM) $M_{\text{LLM}}$, which generates the corresponding answer $M_{\text{LLM}}(\boldsymbol{q};\boldsymbol{o}_j)$ based on the recognized table. 
The reward for each QA pair is computed as the F1-score between the predicted and ground-truth answers, and the overall reward for $\boldsymbol{o}_j$ is the average across all QA pairs for image $\boldsymbol{I}$:
\begin{equation}
	\operatorname{Reward}(\boldsymbol{o}_j)=\frac{1}{|\boldsymbol{QA}|}\sum_{(\boldsymbol{q}, \boldsymbol{a})\in\boldsymbol{QA}}\operatorname{F1}(M_{\text{LLM}}(\boldsymbol{q};\boldsymbol{o}_j),\boldsymbol{a}).
\end{equation}

\noindent\textbf{Filtering-based Stabilization.} Despite its effectiveness, QA-based reward estimation can fail when the TR model generates illegal or repetitive outputs that do not form valid table structures.
Since such outputs cannot yield meaningful answers, they are assigned a reward of zero. 
However, naively including these samples introduces reward noise, as they inflate the relative advantage of valid responses and destabilize training.
To address this issue, we apply illegal-sample filtering, which discards invalid recognition results with zero rewards during GRPO training. 
As shown in Section~\ref{sec:exp_ablation_study}, this strategy significantly stabilizes QA-based rewards, resulting in smoother and more reliable GRPO.

\subsection{Adaptive Data Engine}\label{sec:data_engine}
Building upon the above formulation, we next present how \opt adaptively constructs supervisory signals tailored for the VLM to be fine-tuned.
This includes selecting unlabeled samples that yield the most significant optimization signals (Section~\ref{sec:sampling}) and generating QA pairs that provide richer feedback (Section~\ref{sec:qa_generation}).

\subsubsection{Response-Consistency Sampling}\label{sec:sampling}
Learning from unlabeled data allows one to scale up the training set to cover real-world tables with diverse layouts. However, not all samples contribute equally to learning, and identifying which ones offer the greatest training value for the VLM is non-trivial. 
For example, clustering-based methods~\cite{li2025eagle} can measure sample diversity, but they cannot determine which samples are most beneficial for improving the model to be fine-tuned.
Alternatively, human-in-the-loop approaches~\cite{chen2025logics} can effectively mine hard cases but are subjective and impractical for large-scale datasets.

Since GRPO benefits from samples that elicit diverse model responses, table images that cause the TR model to produce varied recognition outputs are particularly valuable. To capture this, we sample multiple recognition results for each table image and measure their pairwise structural similarity using the Tree Edit Distance-based Similarity (TEDS) metric~\cite{zhong2020image}.
Formally, for each image $\boldsymbol{I}$, the TR model $M_{\text{TR}}$ generates a set of $K$ individual responses $\{\boldsymbol{o}_i\}_{i=1}^{K}$.
The consistency score for image $\boldsymbol{I}$ is then defined based on the aggregated pairwise TEDS similarity among these responses:
\begin{equation}
\operatorname{Consistency}(\boldsymbol{I}) = \frac{2}{K^2-K}\sum_{1\leq i < j \leq K} \text{TEDS}(\boldsymbol{o}_i,\boldsymbol{o}_j).
\end{equation}
A lower consistency score indicates higher response diversity and greater model uncertainty, making the sample more valuable for GRPO training.

In principle, response-consistency sampling can be performed online by re-evaluating samples throughout training. 
However, this approach is computationally prohibitive.
We found that conducting this sampling offline already provides stable and effective improvements, as shown in Section~\ref{sec:exp}.

\subsubsection{Attention-based Diverse QA Generation}\label{sec:qa_generation}
QA pairs are the driving force of \opt to learn from unlabeled table images. 
They must satisfy two key requirements: (i) the QA set for each image should cover different table regions to enrich learning signals, and (ii) their correctness must be verifiable. 
A straightforward approach is to prompt a VLM to generate QA pairs. 
However, as shown in Figure~\ref{fig:attn_qa}, single-pass generation often covers only part of the table, while multi-pass generation tends to introduce paraphrased QA pairs that are sourced from overlapping areas. 
Consequently, even if the TR model produces diverse recognition outputs, the QA-based rewards may remain limited, hindering the effectiveness of GRPO training.

\begin{figure}
    \centering
    \includegraphics[width=\linewidth]{figs/attn_qa.pdf}
    \vspace{-5mm}
    \caption{
    Single-time QA generation captures limited table content, while multiple samplings introduce redundant or overlapping QA pairs.
    The proposed attention-guided QA generation leverages attention distributions to diversify question sources, producing concise and comprehensive QA pairs.
    }
    \label{fig:attn_qa}
    \vspace{-5mm}
\end{figure}

To address this, we leverage the attention mechanism of VLMs, which inherently encodes the visual grounding of textual tokens during answer generation~\cite{kang2025see,baek2025large}. Each answer token attends to visual tokens that provide its evidence, enabling \opt to reason about the visual source of each QA pair. Building on this insight, we design an attention-based QA generation method that explicitly utilizes these attention patterns to identify and filter out QA pairs with limited contributions.

For a given QA pair $(\boldsymbol{q}, \boldsymbol{a})$ generated from image $\boldsymbol{I}$ by the QA generation model $M_{\text{QA}}$, we define its visual source as the set of visual tokens with significant attention weights. 
Specifically, if the table image is tokenized into visual tokens $\boldsymbol{\mathcal{V}}$, the visual source (VS) of the QA pair is defined as:
\begin{equation}
\hspace{-0.69em}\operatorname{VS}((\boldsymbol{q},\boldsymbol{a});\boldsymbol{I},M_{\text{QA}})=\{v \mid \mathcal{A}_{M_{\text{QA}}}(v \mid \boldsymbol{a})>\tau_{\mathcal{A}}, v\in\boldsymbol{\mathcal{V}}\}
\end{equation}
where $\mathcal{A}_{M_\text{QA}}(v \mid \boldsymbol{a})$ denotes the attention score to the visual token $v$ averaged across answer tokens, and $\tau_{\mathcal{A}}$ is a fixed threshold. 
The resulting token list represents the visual grounding of the QA pair, allowing \opt to eliminate pairs that rely on overlapping evidence.

Based on the visual source above, the attention-guided QA generation process involves three steps (Figure~\ref{fig:pipeline}(a)):
\begin{enumerate}
	\item Candidate QA Pool Construction: Given an image $\boldsymbol{I}$, the QA generation (teacher) model $M_{\text{QA}}$ is prompted multiple times to produce a pool of candidate QA pairs: $\operatorname{GenQA}(\boldsymbol{I}) = \{(\boldsymbol{q},\boldsymbol{a})\}$.
	\item Validity Cross-checking: Each candidate pair is cross-checked by an external VLM $M_{\text{Val}}$ to ensure visual dependency and correctness. Specifically, we retain only those pairs that can be correctly answered with the image but not without it: $\operatorname{ValidQA}(\boldsymbol{I})=\{(\boldsymbol{q},\boldsymbol{a})\mid M_{\text{Val}}(\boldsymbol{q};\boldsymbol{I})=\boldsymbol{a}\land M_{\text{Val}}(\boldsymbol{q};\emptyset)\neq a,(\boldsymbol{q},\boldsymbol{a})\in\operatorname{GenQA}(\boldsymbol{I})\}$. 
	\item Attention-Guided QA Selection: We greedily select valid pairs whose visual sources exhibit minimal overlap to ensure broad table coverage. To account for attention sink effects~\cite{kang2025see}, we relax disjointness by an Intersection-over-Union (IOU) threshold between any two QA pairs: $\mathrm{IOU}(\operatorname{VS}((\boldsymbol{q}_i, \boldsymbol{a}_i);\boldsymbol{I},M_{\text{QA}}),\operatorname{VS}((\boldsymbol{q}_j \boldsymbol{a}_j);\boldsymbol{I},M_{\text{QA}}))<\tau_{\mathrm{IOU}}$.
\end{enumerate}
This attention-guided mechanism ensures that QA pairs are both visually grounded and semantically diverse, providing richer and more robust supervision for GRPO training.

\section{\opt-3B}\label{sec:model}
Building upon the above framework, we propose \opt-3B, a model obtained by fine-tuning Qwen2.5-VL-3B-Instruct to surpass the TR performance achievable with existing labeled datasets and establish a new state of the art. 
The overall training pipeline consists of three stages. 
We first leverage labeled datasets to gradually adapt a general-purpose VLM into a strong TR model, establishing the performance limit achievable through supervised learning. 
We then apply \opt to further fine-tune the model in a self-supervised manner using unlabeled data. 
Additional implementation details are provided in the appendix. 
Together with the source code, \opt-3B will be publicly released upon acceptance (now in supplementary materials).

\noindent\textbf{Stage 1: OTSL Tag Warm-up.} Compared to HTML or Markdown representations, OTSL~\cite{lysak2023optimized} provides a compact and well-structured table format. 
It encodes adjacency relationships instead of explicitly predicting colspan and rowspan attributes for merged cells, substantially reducing token length and simplifying structural prediction.
In this first stage, we train the model on large-scale open-source datasets to familiarize it with the OTSL syntax. 
We aggregate data from PubTabNet~\cite{zhong2019image}, SynthTabNet~\cite{nassar2022tableformer}, and MMTab~\cite{zheng2024multimodal}. 
Specifically, we sample 100K instances from each of the four SynthTabNet subsets, 200K from PubTabNet, and 100K from MMTab. 
After removing erroneous samples (e.g., with unclosed HTML tags) and converting all annotations into OTSL, we obtain approximately 700K training instances. 
During this warm-up, only the language model component of Qwen2.5-VL-3B is fine-tuned, while the visual encoder and alignment module remain frozen. 
This allows the model to learn OTSL syntax and structure without disrupting low-level visual representations.

\begin{table*}[t]
	\centering
	\resizebox{\linewidth}{!}{\begin{tabular}{lcccccccccc}
			\toprule
			& \multicolumn{2}{c}{PubTabNet} & \multicolumn{2}{c}{OmniDocBench} & \multicolumn{2}c{CC-OCR} & \multicolumn{2}{c}{OCRBench} & \multicolumn{2}{c}{Overall} \\
			\cmidrule(lr){2-3} \cmidrule(lr){4-5} \cmidrule(lr){6-7} \cmidrule(lr){8-9} \cmidrule(lr){10-11} 
			& TEDS & S-TEDS & TEDS & S-TEDS & TEDS & S-TEDS & TEDS & S-TEDS & TEDS & S-TEDS \\
			\midrule
			\multicolumn{9}{l}{Expert TR models} \\
			\midrule
			SLANNet-plus & 86.57 & \textbf{96.43} & 81.90 & 89.08 & 50.93 & 65.84 & 65.55 & 77.73 & 68.19 & 79.21 \\
			UniTable & 86.44 & \underline{95.66} & 82.76 & 89.82 & 57.84 & 70.47 & 67.73 & 78.65 & 70.86 & 80.81 \\
			\midrule
			\multicolumn{9}{l}{General-purpose VLMs} \\
			\midrule
			InternVL3.5-241B-A30B & 83.75 & 88.76 & 86.03 & 90.53 & 62.87 & 69.52 & 79.50 & 85.81 & 78.41 & 84.18 \\
			Qwen2.5-VL-72B & 84.39 & 87.91 & 87.85 & 91.80 & 81.22 & 86.48 & 81.33 & 86.58 & 83.52 & 88.33 \\
			Qwen3-VL-235B-A22B & - & - & 91.02 & \underline{94.97} & 80.98 & 86.19 & 84.12 & 88.15 & 85.83 & 90.07 \\
			Gemini~2.5~Pro & - & - & 90.90 & 94.32 & \textbf{85.56} & \underline{90.07} & 88.94 & 89.47 & \underline{88.93} & \underline{91.23} \\
			GPT-4o & 76.53 & 86.16 & 78.27 & 84.56 & 66.98 & 79.04 & 70.51 & 79.55 & 72.44 & 81.15 \\
			GPT-5  & - & -         & 84.91 & 89.91 & 63.25 & 74.09 & 79.91 & 88.69 & 78.30 & 86.21 \\
			\midrule
			\multicolumn{9}{l}{Document-parsing VLMs} \\
			\midrule
			dots.ocr & 90.65 & 93.76 & 88.62 & 92.86 & 75.42 & 81.65 & 82.04 & 86.27 & 82.95 & 87.58 \\
			DeepSeek-OCR & - & - & 83.79 & 87.86 & 68.95 & 75.22 & 82.64 & 87.33 & 80.31 & 85.11 \\
			PaddleOCR-VL & - & - & \underline{91.12} & 94.62 & 79.62 & 85.04 & 79.29 & 83.93 & 83.36 & 87.77 \\
			MinerU2.5 & 89.07 & 93.11 & 90.85 & 94.68 & 79.76 & 85.16 & \underline{87.13} & \underline{90.62} & 86.82 & 90.81 \\
			\midrule
			\opt-3B & \textbf{91.79} & 93.81 & \textbf{91.60} & \textbf{95.01} & \underline{84.90} & \textbf{90.17} & \textbf{90.76} & \textbf{94.03} & \textbf{89.88} & \textbf{93.60} \\
			\bottomrule
	\end{tabular}}
    \vspace{-3mm}
	\caption{\opt-3B achieves consistently high TR performance, in TEDS and S-TEDS, across four benchmarks, which is unattainable by 12 existing methods that include expert TR models, general-purpose VLMs, and those fine-tuned for document parsing.}
    \vspace{-3mm}
	\label{tab:exp_general_performance}
\end{table*}

\noindent\textbf{Stage 2: Supervised Fine-Tuning.} In the second stage, we improve robustness and generalization by fine-tuning the model on real-world table images collected from open-source datasets~\cite{zhang2024semv2,yang2023large,long2021parsing} and web resources~\cite{talocrtable}. 
We curate approximately 50K samples and fine-tune the model with all parameters unfrozen. 
This stage bridges the gap between synthetic and real-world layouts, reaching the performance limit achievable through supervised learning.

\noindent\textbf{Stage 3: \opt.} In the final stage, we apply \opt to further optimize the model through GRPO with QA-based rewards. 
Following the data curation procedure described in Section~\ref{sec:data_engine}, we construct the RL training dataset as follows. 
We collect PDFs from web sources, use DocLayout-YOLO~\cite{zhao2024doclayout} for layout detection, and crop tables to form an unlabeled table image pool of approximately 100K samples.
Using the Stage-2 TR model, we perform response-consistency sampling to identify images that yield diverse recognition outputs. 
We observe that samples with consistency scores below 0.4 are often noisy or non-tabular. 
Therefore, we uniformly sample images with scores in the range of 0.4–1.0, at an interval of 0.1 to ensure diversity across uncertainty levels, retaining about 50K informative samples for RL training. 
For each selected image, we generate QA pairs using Qwen2.5-VL-72B-Instruct as $M_\text{QA}$ and empirically select layer 72 for attention computation with attention thresholds $\tau_{\mathcal{A}}=0.01$ and $\tau_{\mathrm{IOU}}=0.3$ (Section~\ref{sec:qa_generation}). 
After validity cross-checking by InternVL3-78B, we retain roughly 30 diverse QA pairs per image, resulting in a balanced and information-rich QA dataset. 
During GRPO training (Section~\ref{sec:rl_training}), Qwen3-8B acts as $M_\text{LLM}$, responsible for answering the generated QA pairs based on the recognized table outputs. 


\section{Experimental Evaluation}\label{sec:exp}

We conduct empirical studies to evaluate and analyze the effectiveness of the proposed \opt framework and the resulting \opt-3B model.

\noindent\textbf{TR Benchmarks.} We evaluate \opt-3B across three widely adopted benchmarks: OmniDocBench~\cite{ouyang2025omnidocbench}, CC-OCR~\cite{yang2025cc}, and OCRBench v2~\cite{fu2024ocrbench}.
OmniDocBench consists of digital PDFs with a wide range of table types.
We use the latest version, OmniDocBench v1.5, which includes more complex tables and crops all table images based on the provided layout annotation, resulting in 512 samples for evaluation.
CC-OCR consists of 300 scanned and photographed table images, including challenging samples such as long tables, complex structures, and handwritten samples.
OCRBench v2 is the latest version of OCRBench.
We employ its table parsing subset, which includes 700 table images, featuring diverse real-world table images with a variety of layouts and visual complexities.

\noindent\textbf{Comparison Schemes.} We compare \opt-3B with three categories of models: (1) Expert TR models: SLANNet-plus~\cite{cui2025paddleocr} and UniTable~\cite{peng2024unitable}.
(2) General-purpose VLMs: InternVL3.5-241B-A30B~\cite{wang2025internvl3}, Qwen2.5-VL-72B, Qwen3-VL-235B-A22B~\cite{bai2025qwen2}, Gemini~2.5~Pro~\cite{comanici2025gemini}, GPT-4o~\cite{hurst2024gpt}, and GPT-5~\cite{gpt5}. These models are adopted for TR using the prompt provided in the appendix.
(3) Document-parsing VLMs: dots.ocr~\cite{dotsocr}, DeepSeek-OCR~\cite{wei2025deepseek}, PaddleOCR-VL~\cite{cui2025paddleocrvl}, and MinerU2.5~\cite{niu2025mineru2}, each evaluated with their default prompts.

\noindent\textbf{Outline.} We first examine how \opt-3B advances a wide range of state-of-the-art models across different benchmarks in Section~\ref{sec:exp_general_benchmarks}. Then, we analyze the enablers of such advancements via ablation studies in Section~\ref{sec:exp_ablation_study}. Finally, we discuss the broader utility for \opt as a scalable data annotator in Section~\ref{sec:exp_label_annotation}.

%

\subsection{Advancing TR Performance}
\label{sec:exp_general_benchmarks}
Table~\ref{tab:exp_general_performance} summarizes the TR performance of \opt-3B and 12 baselines over four datasets. We report TEDS~\cite{huang2023improving} and S-TEDS (structure-only TEDS) to assess both holistic and structural accuracy.

\noindent\textbf{Expert TR Models.} \opt-3B achieves substantially better TR on diverse scenarios, except for PubTabNet in S-TEDS, due to dataset-specific overfitting of expert models. Specifically, \opt-3B surpasses UniTable by 27.06 and 23.03 TEDS on CC-OCR and OCRBench v2, respectively, confirming its robustness to complex layouts. 

\noindent\textbf{General-purpose VLMs.} Despite being trained with significantly fewer parameters and data, \opt-3B still outperforms all these large-scale general-purpose VLMs on most benchmarks. In particular, Qwen2.5-VL-72B has been used as the QA generation (teacher) model (i.e., $M_{\text{QA}}$ in Section~\ref{sec:qa_generation}) to generate table QAs for fine-tuning \opt-3B. We observe a 6.36 TEDS improvement overall, indicating that \opt successfully extracts and refines useful supervision beyond direct distillation. Another noteworthy baseline is Gemini 2.5 Pro, the proprietary model built with massive human and computational resources. It has incorporated table recognition data during pretraining and hence achieved the most competitive performance (88.93 TEDS and 91.23 S-TEDS). Still, \opt-3B consistently outperforms it in most benchmarks, except for CC-OCR (85.56 TEDS vs 84.90 TEDS).

\noindent\textbf{Document-parsing VLMs.} \opt-3B exhibits better performance than existing models tailored for document parsing in all benchmarks. 
Although its performance is marginally better than the earlier SOTA model, PaddleOCR-VL, on OmniDocBench, which primarily consists of digital table images, \opt-3B outperforms on the CC-OCR and OCRBench benchmarks by a large margin (5.28 and 11.47 TEDS).
The advantages can also be observed when compared with MinerU2.5, which has been trained with data several orders of magnitude more than \opt-3B, with manual annotation and distillation from Gemini 2.5 Pro. \opt-3B can still offer better TR in all cases. 
These results validate that \opt not only enhances performance but also improves robustness to diverse real-world distributions.

\begin{table}[t]
	\centering
	\resizebox{\linewidth}{!}{\begin{tabular}{lcccc}
			\toprule
			& \begin{tabular}[c]{@{}c@{}}OmniDoc\\Bench\end{tabular}  & \begin{tabular}[c]{@{}c@{}}CC\\OCR\end{tabular} & \begin{tabular}[c]{@{}c@{}}OCR\\Bench\end{tabular} & Overall \\
			\midrule
			Stage-1 & 87.65 & 72.28 & 73.06 & 77.85 \\
			\midrule
			Stage-2 & 90.08 & 82.48 & 90.08 & 88.57 \\
			+\begin{tabular}[c]{@{}l@{}}Qwen2.5-VL-72B\\\&SFT\end{tabular} & \begin{tabular}[c]{@{}c@{}}84.41 \\(\redtext{-5.67})\end{tabular} & \begin{tabular}[c]{@{}c@{}}70.54 \\(\redtext{-11.94})\end{tabular} & \begin{tabular}[c]{@{}c@{}}80.87 \\(\redtext{-9.21})\end{tabular} & \begin{tabular}[c]{@{}c@{}}80.02 \\(\redtext{-8.53})\end{tabular} \\
			+\begin{tabular}[c]{@{}l@{}}Qwen2.5-VL-72B\\\&GRPO\end{tabular}  & \begin{tabular}[c]{@{}c@{}}86.19 \\(\redtext{-3.89})\end{tabular} & \begin{tabular}[c]{@{}c@{}}78.12\\(\redtext{-4.36})\end{tabular} & \begin{tabular}[c]{@{}c@{}}84.16\\(\redtext{-5.92})\end{tabular} & \begin{tabular}[c]{@{}c@{}}83.65\\(\redtext{-4.92})\end{tabular} \\
			\midrule
			\opt-3B & \textbf{91.60} & \textbf{84.90} & \textbf{90.76} & \textbf{89.88} \\
			\bottomrule
	\end{tabular}}
	\caption{Using the QA generation model (Qwen2.5-VL-72B) to generate pseudo-labels for SFT or GRPO could not achieve high TR performance, measured in TEDS. The table QA-based approach in \opt can mitigate the impact caused by imperfect annotations.}
	\label{tab:exp_ablation_study}
\end{table}

\subsection{Ablation Studies}
\label{sec:exp_ablation_study}
\opt-3B exhibits consistent improvements over existing methods across different benchmarks. Next, we analyze the enablers of reaching such a TR capability.

\begin{figure}[t]
	\centering
	\includegraphics[width=\linewidth]{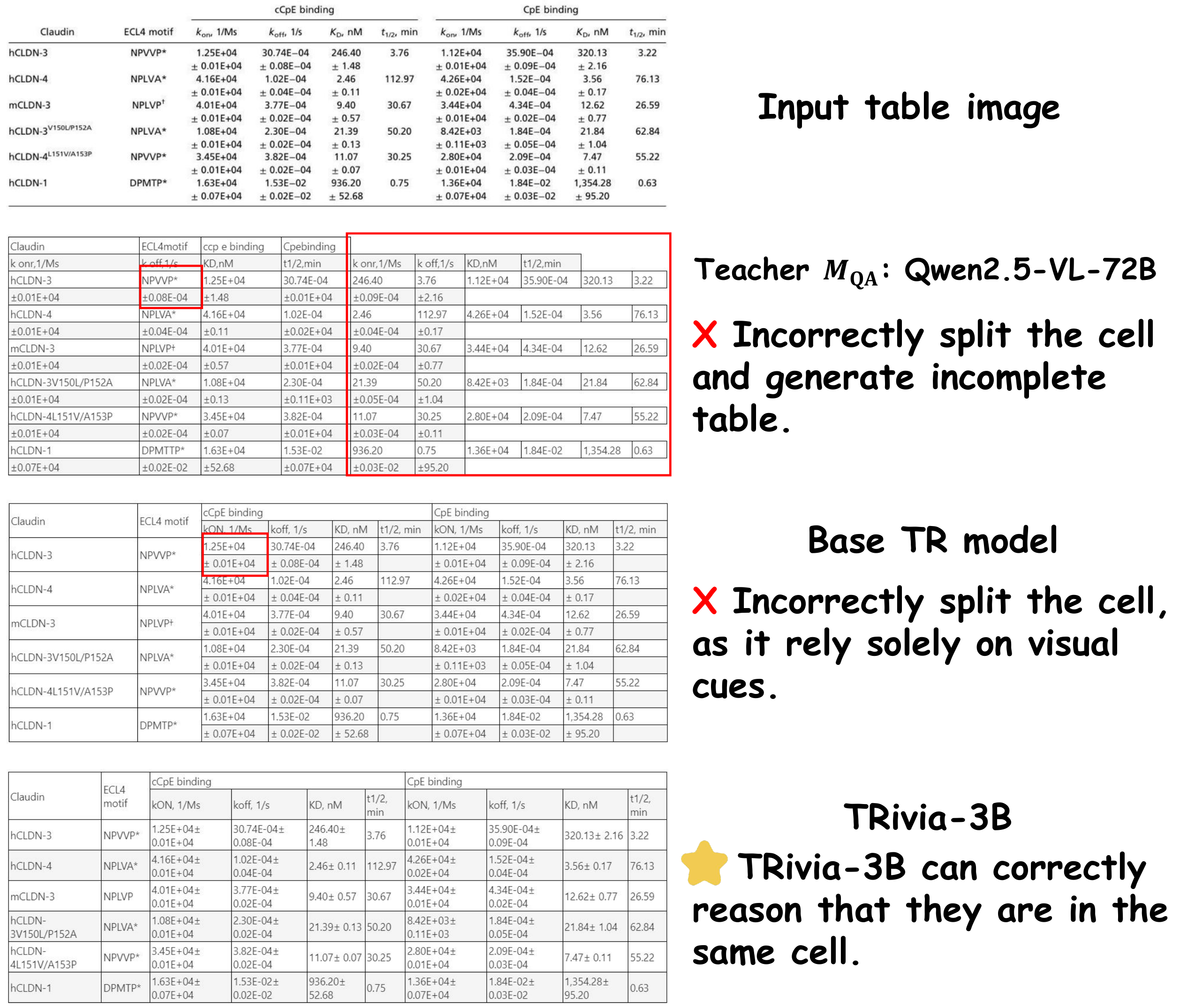}
	\caption{
		The teacher model $M_\text{QA}$ used for QA generation could not directly generate correct annotations, but is sufficient to create QAs with \opt to fine-tune the base model and lead to \opt-3B that can handle complex structures.
	}
	\label{fig:case_1}
\end{figure}

\noindent\textbf{Table QA-based Supervisory Signals.} Our table QA-based proxy reward allows \opt to learn from a VLM with imperfect annotations. To understand this improvement, we use the QA generation model, Qwen2.5-VL-72B, to (i) produce HTML-based table recognition on training images, (ii) convert them into OTSL format, and (iii) fine-tune the Stage-2 VLM with these pseudo-labels using either supervised fine-tuning (SFT) or GRPO. Qwen2.5-VL-72B cannot produce high-quality recognition results, as shown in Figure~\ref{fig:case_1}, depicting a training sample that the model fails to split the cells and generate the table correctly. 

Based on these imperfect labels, our results reported in Table~\ref{tab:exp_ablation_study} reveal that the fine-tuned models have significantly worsened TR performance. 

SFT leads to an average decrease of 8.37 TEDS, even underperforming Stage-1 on OmniDocBench, while GRPO mitigates the decline slightly but still lags by 4.92 TEDS. We attribute this degradation to biased or semantically inconsistent annotations produced by Qwen2.5-VL-72B in the target domain. In contrast, the \opt framework produces high-quality QA pairs to avoid such bias, producing consistent supervision signals and yielding better results.

%
%
%
%
%

\begin{figure}[t]
	\centering
    \vspace{-3mm}
	\includegraphics[width=\linewidth]{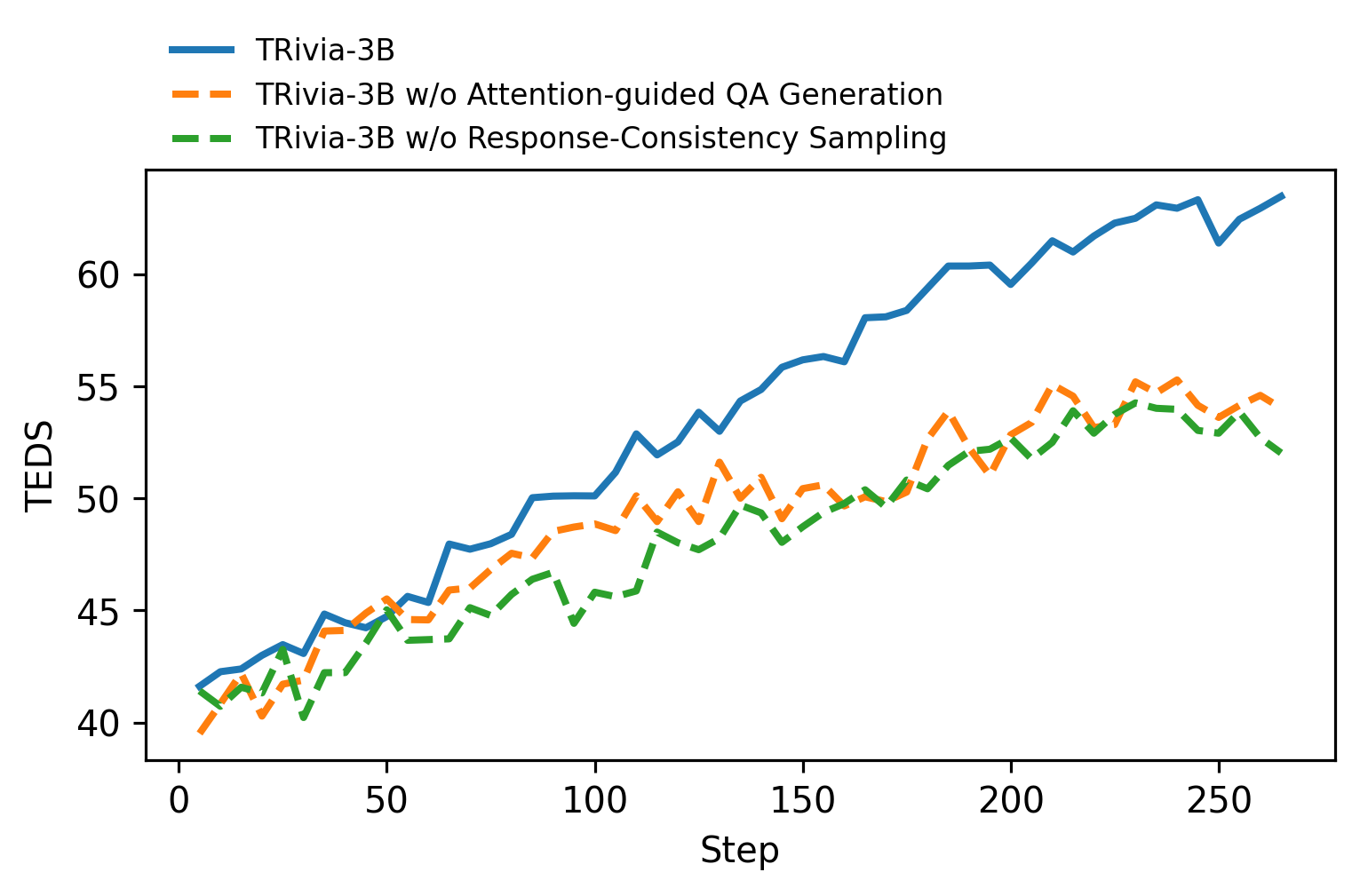}
    \vspace{-6mm}
	\caption{\opt-3B benefits significantly from the diverse QAs generated by the attention-guided mechanism and the training samples that yield diverse outputs.
	}
    \vspace{-3mm}
	\label{fig:in_domain_validation}
\end{figure}

\noindent\textbf{Attention-guided QA Generation.}
Table QA-based rewards are naturally sparse, as each QA pair assesses only a limited table region.
Our proposed attention-guided QA generation enriches supervision diversity by widening question sources according to attention distributions.
To show the advantage of such a design, we curate a test set of 80 table samples of varying levels of complexity and create annotations manually for evaluation purposes.
As shown in Figure~\ref{fig:in_domain_validation} (orange line), removing attention-guided QA generation leads to a significant drop in TR performance. Our investigation found that the fine-tuned model is particularly weak in handling structurally complex or visually ambiguous tables. 

\noindent\textbf{Response-Consistency Sampling.}
To prioritize informative samples, \opt adopts response-consistency sampling, which favors table images yielding diverse model outputs.
This design aligns with the GRPO principle of optimizing relative advantages within a response group.
Compared to random sampling, response-consistency sampling accelerates convergence and boosts TEDS from 52.0 to 63.5, as shown in Figure~\ref{fig:in_domain_validation} (green line).
Although the sampling process occurs offline, it consistently enhances optimization efficiency, confirming that response-consistency sampling effectively identifies informative and challenging examples for optimization.

\begin{figure}[t]
	\centering
	\includegraphics[width=0.93\linewidth]{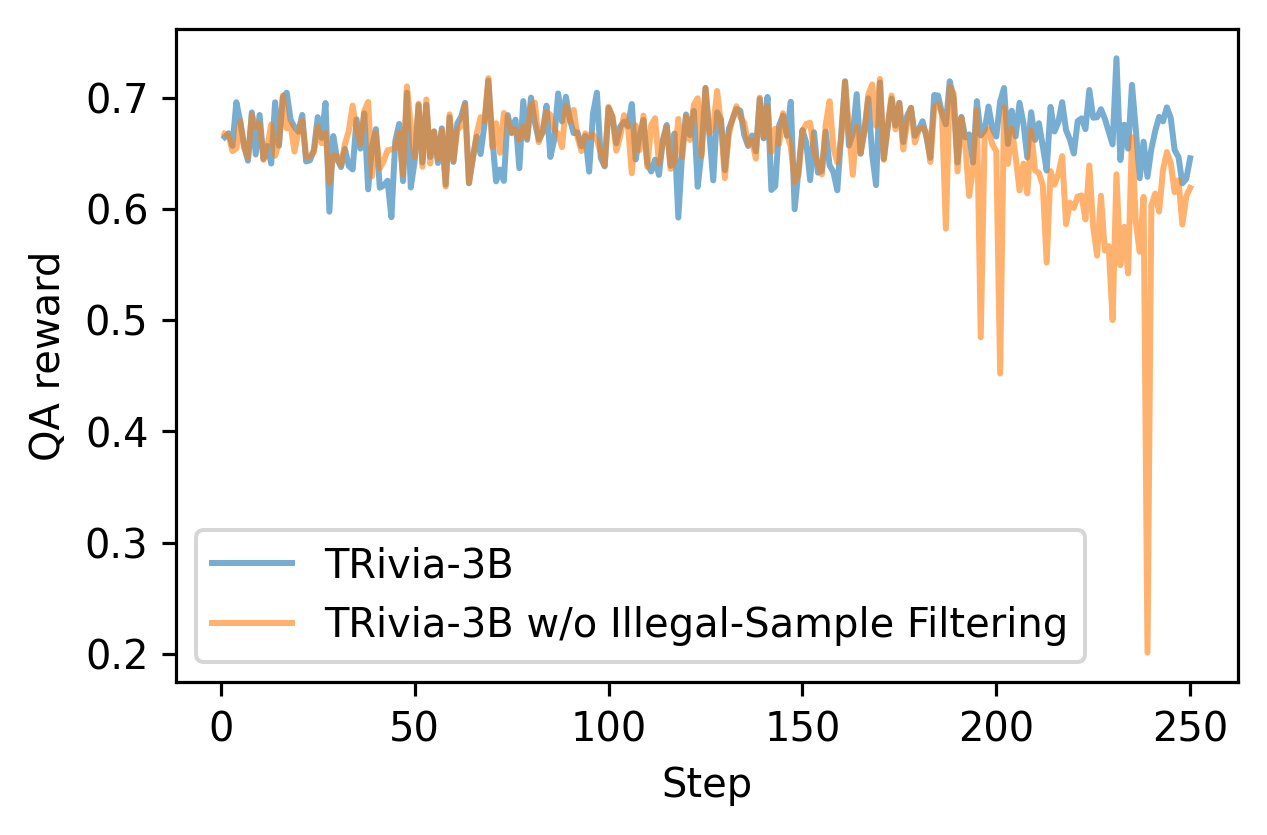}
    \vspace{-4mm}
	\caption{
		Illegal-sample filtering is crucial for stabilizing the fine-tuning of the TR model, as it suppresses reward noise caused by invalid responses.
	}
    \vspace{-4mm}
	\label{fig:exp_sample_filter}
\end{figure}

\noindent\textbf{Illegal-Sample Filtering.}
Table QA-based rewards depend on valid table recognition outputs.
When the model generates invalid responses, the LLM cannot produce correct answers, forcing these samples to receive zero reward.
This artificially compresses the reward distribution and destabilizes GRPO.
We mitigate this issue through illegal-sample filtering, which removes invalid responses before advantage computation.
As shown in Figure~\ref{fig:exp_sample_filter}, without filtering (orange line), the learning progressively becomes unstable and eventually leads to significant fluctuation in QA rewards. 
Instead, illegal-sample filtering (blue line) successfully stabilizes the learning.
As a result, this approach reduces the convergence step by approximately 25\% and improves final performance on test set by 3 TEDS.
\subsection{\opt as a Data Annotator}
\label{sec:exp_label_annotation}
Beyond achieving state-of-the-art performance in table recognition, another crucial requirement is to transfer this capability to other models, such as distilling knowledge into smaller, more efficient architectures or enhancing the OCR abilities of general-purpose VLMs.
To this end, we demonstrate the broader potential of \opt as an automated data annotation system capable of generating reliable pseudo-labels for unlabeled data.
Specifically, we use \opt-3B to create pseudo-labels for a set of unlabeled table images it did not see during its fine-tuning process. Then, we employ standard SFT to optimize a Stage-2 model. 
%
As shown in Table~\ref{tab:distillation}, the distilled model (SFT w/ \opt-3B labels) achieves nearly identical performance to \opt-3B across all benchmarks, confirming the high fidelity of generated annotations.
\begin{table}[t]
	\centering
	\resizebox{\linewidth}{!}{\begin{tabular}{lcccc}
			\toprule
			& \begin{tabular}[c]{@{}c@{}}OmniDoc\\Bench\end{tabular}  & \begin{tabular}[c]{@{}c@{}}CC\\OCR\end{tabular} & \begin{tabular}[c]{@{}c@{}}OCR\\Bench\end{tabular} & Overall \\
			\midrule
			Stage-2 & 90.08 & 82.48 & 90.08 & 88.57 \\
			+ \begin{tabular}[c]{@{}l@{}}SFT w/ \\\opt-3B labels\end{tabular} & 91.37 & \textbf{85.84} & 90.75 & 89.99 \\
			\midrule
			\opt-3B & \textbf{91.60} & {84.90} & \textbf{90.76} & \textbf{89.88} \\
			\bottomrule
	\end{tabular}}
	\caption{
		The distilled model (SFT w/ \opt-3B labels) achieves comparable performance to \opt-3B, confirming the reliability of \opt-based annotations.
	}
	\label{tab:distillation}
\end{table}
Moreover, on the challenging CC-OCR benchmark, which is characterized by visually complex and layout-diverse tables, the distilled model even slightly outperforms \opt-3B.
Notably, this outcome contrasts with the poor performance obtained when distilling from Qwen2.5-VL-72B (see Table~\ref{tab:exp_ablation_study}), which fails to yield accurate HTML annotations for the same data.
In summary, \opt establishes a new paradigm for table data annotation, enabling dynamic model adaptation to dataset characteristics rather than relying on a fixed annotator.
These findings further highlight the potential of \opt as a scalable, fully automated alternative to manual labeling or costly proprietary model distillation for constructing high-quality TR datasets at scale.



\section{Conclusions}
In this paper, we introduce \opt, a self-supervised fine-tuning framework that enables VLMs to learn table recognition directly from unlabeled table images.
Built upon \opt, \opt-3B establishes a new state-of-the-art among TR models, outperforming both specialized and proprietary models across multiple benchmarks.
Beyond improving recognition performance, \opt also serves as a scalable annotation engine, generating pseudo-labels that rival human and proprietary annotations.
This work opens promising directions for self-supervised document parsing. 
Its annotation-free design provides a foundation for extending similar principles to broader multimodal tasks for learning at scale.

\section{Acknowledgments}
This research is partially supported by the RGC Early Career Scheme (Project \#27211524) and Croucher Start-up Allowance (Project \#2499102828).
Any opinions, findings, or conclusions expressed in this material are those of the authors and do not necessarily reflect the views of RGC and the Croucher Foundation.
This research is partially supported by Shanghai Artificial Intelligence Laboratory.

{
    \small
    \bibliographystyle{ieeenat_fullname}
    \bibliography{main}
}

\clearpage
\setcounter{page}{1}
\maketitlesupplementary
\appendix

\newcommand{\tb}[1]{\textbf{#1}}
\newcommand{\ud}[1]{\underline{#1}}

\section{More Training Details}
\label{supp_sec:training_details}
\begin{table}[!hbp]
    \centering
    \resizebox{\linewidth}{!}
    { 
    \begin{tabular}{@{}ll|c|c|c@{}}
    \toprule
    & & \textbf{Stage-1} & \textbf{Stage-2} & \textbf{Stage-3} \\
    & & & & \\
    \midrule 
    \multirow{2}{*}{\rotatebox[origin=c]{90}{\footnotesize \textit{Vision}}}
    & \textbf{Max Resolution}   
        & $1280 \times 28 \times 28$
        & $1280 \times 28 \times 28$
        & $1280 \times 28 \times 28$ \\
    & \#Tokens per Image 
        & $256 \sim 1280$ 
        & $256 \sim 1280$ 
        & $256 \sim 1280$  \\
    \midrule 
    \multirow{2}{*}{\rotatebox[origin=c]{90}{\footnotesize \textit{Data}}}
    & \textbf{Dataset} 
        & Synthetic Data
        & Real-world Data
        & \opt Curated Data \\
    & \#Samples 
        & 700K 
        & 50K 
        & 50K \\
    \midrule
    \multirow{2}{*}{\rotatebox[origin=c]{90}{\footnotesize \textit{Model}}}
    & \textbf{Trainable} 
        & LLM
        & All 
        & All \\
    & \textbf{Sequence Length}
        & 8192 
        & 8192 
        & 8192 \\
    \midrule 
    \multirow{3}{*}{\rotatebox[origin=c]{90}{\footnotesize \textit{Training}}}
    & \textbf{Batch Size} 
        & 32 
        & 32 
        & 128 \\
    & \textbf{LR: $\psi_{\text{ViT}}$} 
        & 2 $\times 10^{-6}$ 
        & 2 $\times 10^{-6}$ 
        & 2 $\times 10^{-7}$ \\    
    & \textbf{LR: $\{\theta_{\text{MLP}}, \phi_{\text{LM}}\}$} 
        & 1 $\times 10^{-5}$ 
        & 1 $\times 10^{-5}$ 
        & 1 $\times 10^{-6}$ \\
    & \textbf{Epoch} 
        & 1 
        & 2 
        & 1 \\
    \bottomrule
    \end{tabular}
    } 
    \caption{Training setup and hyperparameters in three training stages.} 
    \label{tab:training_strategy}
\end{table}
The training configurations for the three stages are summarized in Table~\ref{tab:training_strategy}. 
We use Qwen2.5-VL-3B as the backbone model.
Across all stages, the number of image tokens ranges from $256$ to $1280$, corresponding to image resolutions from $256 * 28 * 28$ to $1280 * 28 * 28$. 
The prompt templates used for table recognition are provided in Table~\ref{tab:tab_recog_prompt}.
In the third stage, which incorporates GRPO training, the sampling temperature is set to 1.2.
We generate $G=16$ samples per step and use a constant learning rate scheduler.
All experiments are conducted with 8 X A100 80GB GPUs. 
Stage 1 training requires approximately one day, Stage 2 about two hours, and Stage 3 (GRPO) around two days.

\section{Dataset Construction}
For stage 1 and 2, we removes all samples with incomplete HTML tags during dataset construction.
In stage 2, when only table structure tags were available~\cite{zhang2024semv2,long2021parsing}, we employ Qwen2.5-VL-72B to perform OCR over each cell’s bounding box to recover textual content.

In stage 3, we begin with 100K unlabeled images.
We employ response-consistency sampling using the stage-2 model with a temperature of 1.0, producing eight outputs per image.
The consistency score is computed via pairwise TEDS among these outputs.
Then, we calculate the consistency score using pairwise TEDS among the 8 outputs. 
Images are uniformly sampled across consistency score intervals from 0.4 to 1.0 with a step size of 0.1.
To promote diversity, we ensure that all table images originate from distinct PDFs. 
The filtering results in 50K selected images.

We then generate QA pairs using the attention-guided QA generation for these 50K images. 
We use Qwen2.5-VL-72B as $M_{\text{QA}}$, generating 16 QAs per image with a temperature of 1.0 and prompt as shown in Table~\ref{tab:qa_gen_prompt}.
Invalid JSON outputs are discarded.
For cross-checking filtering, we adopt InternVL3-78B as $M_{\text{Val}}$ to answer each question with and without table images as input and not having table images as input. 
We retain QA pairs whose F1 score exceeds 0.9 with the image but falls below 0.3 without it. 
Furthermore, we use $M_{\text{QA}}$ to obtain the visual source of each QA pair and greedily search the final sets of QA pairs such that the IOU between any two QA pairs is less than 0.3.
We remove images with fewer than 3 valid QAs to maintain reliable QA reward estimation. 
The final dataset comprises 48,470 images, each associated with an average of 28.3 QAs.

\section{More Details about Experiments}
All experiments are conducted using eight A100 80GB GPUs. 
For inference, we employ vLLM by default for all compatible models.

\noindent\textbf{General-purpose VLMs for TR.}
When using general-purpose VLMs for table recognition, we perform prompt optimization to achieve the best results. 
We compare various templates from benchmarks such as OmnidocBench, CC-OCR, and OCRBench v2, and find that a unified prompt design (shown in Table~\ref{tab:tab_recog_prompt}) achieves the best overall performance.
During generation, we set the sampling temperature to 0.2, which reduces repetitions relative to temperature 0 and mitigates hallucinations compared to temperature 1.0, yielding around a 3\% performance gain. 
For Gemini 2.5 Pro, we enable “thinking mode”, which improves performance by approximately 3\%.
For the Qwen2.5-VL and Qwen3-VL, we set the number of image tokens to $256 \sim 1280$, which is the same as \opt-3B.

\noindent\textbf{Document parsing VLMs for TR.}
For document parsing VLMs, we adopt their specialized table recognition prompts.
In the case of PaddleOCR-VL, we disable unwrapping and document orientation classification modules, as disabling them empirically improves performance.

\section{More experimental results}

\subsection{Extension to Multi-Task OCR Learning}
\label{supp_sec:multitask_extension}

\begin{table}[!h]
    \centering
    \vspace{-0.5em}
    \begin{minipage}[b]{.5\linewidth}\setlength\tabcolsep{2pt}\footnotesize
        \centering
        \begin{tabular}{|l|c|c|c|c|}
            \hline
            & \textbf{\begin{tabular}[c]{@{}c@{}}(a)\end{tabular}} & \textbf{\begin{tabular}[c]{@{}c@{}}(b)\end{tabular}} & \textbf{\begin{tabular}[c]{@{}c@{}}(c)\end{tabular}} & \textbf{\begin{tabular}[c]{@{}c@{}}(d)\end{tabular}}  \\ \hline
            Base   &      80.73           &   31.52              &    70.01        &    56.73    \\ \hline
            Stage 2      &      88.84           &   40.19              &    80.11        &    60.55        \\ \hline
            TRivia       & \textbf{90.20}       & \textbf{61.95}       & \textbf{86.10}  & \textbf{72.96}      \\ \hline
        \end{tabular}\vspace{-0.6em}
        \caption{Complex Table Types}\label{tab:SEei-1}
    \end{minipage}~
    \begin{minipage}[b]{.5\linewidth}\setlength\tabcolsep{3pt}\footnotesize
        \centering
        \begin{tabular}{|l|c|c|}
            \hline
            & \textbf{TR}   & \textbf{KIE} \\ \hline
            Base      & 80.24   & 76.06             \\ \hline
            TRivia: TR       & 82.07   & 76.08          \\ \hline
            TRivia: TR+KIE   & 81.95   & 91.01    \\ \hline
        \end{tabular}\vspace{-0.6em}
        \caption{Multiple OCR tasks}\label{tab:SEei-2}
    \end{minipage}\vspace{-1em}
\end{table}

As \opt is defined through task-specific QA rewards, it can in principle be generalized to other OCR and document understanding tasks.
In this section, we provide a preliminary study on extending \opt beyond table recognition. 
As a first step, we study a joint table recognition (TR) and key information extraction (KIE) setting using a small-scale mixed training setup.
For table recognition, we sample 3K data from our stage 3 data and use the overall TEDS on three benchmarks.
For KIE, we employ CORD~\cite{park2019cord} for training and evaluation.
The results are shown in Table~\ref{tab:SEei-2}.

Applying \opt only to the TR component already improves table recognition performance while maintaining KIE accuracy. 
When QA rewards are further introduced for both TR and KIE, the model achieves a substantial gain on KIE while preserving TR performance. 
Although a comprehensive study on additional tasks such as text recognition, formula recognition, and layout analysis is beyond the scope of this work, these results provide initial evidence that TRivia is compatible with multi-task OCR learning and can serve as a general self-supervised adaptation strategy.

\subsection{Performance on Complex Table Types}
\label{supp_sec:complex_tables}

To better understand where \opt brings the largest gains, we evaluate it on several representative complex table scenarios covered by OmniDocBench, CC-OCR, and OCRBench, including tables containing formulas, rotated tables, large tables, and borderless tables with merged cells. 
The results are summarized in Table~\ref{tab:SEei-1}.

\opt improves performance consistently across all four settings. 
The most significant gain is observed on rotated tables, where the improvement over the base model reaches +21.76 TEDS. 
Large gains are also obtained on large tables and borderless merged-cell tables. 
These results indicate that the proposed self-supervised training strategy is especially effective for visually and structurally challenging cases that are underrepresented in conventional synthetic or benchmark-style training data.

\begin{figure}[!h]
    \centering\vspace{-1.2em}
    \begin{minipage}{.5\linewidth}
        \centering
        \includegraphics[width=0.85\linewidth]{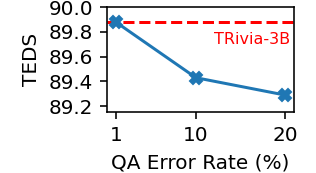}\vspace{-0.7em}
        \caption{QA Error Impact}\label{fig:x3HF-1}
    \end{minipage}~
    \begin{minipage}{.5\linewidth}
        \centering
        \includegraphics[width=0.85\linewidth]{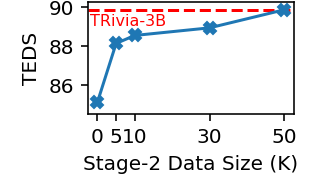}\vspace{-0.7em}
        \caption{Stage-2 Data Size}\label{fig:x3HF-2}
    \end{minipage}\vspace{-1.2em}
\end{figure}

\subsection{Robustness to QA Noise}
\label{supp_sec:qa_noise}

An important practical question is whether the effectiveness of \opt depends critically on the correctness of the automatically generated QA pairs. 
To evaluate this, we conduct an additional robustness study by artificially injecting incorrect QA pairs into the training set. 
The results are shown in Figure~\ref{fig:x3HF-1}. 
We observe that \opt is relatively insensitive to moderate QA corruption: even when $20\%$ of the QA pairs are intentionally replaced with incorrect ones, the degradation in table recognition performance remains limited.

This behavior can be explained by the reward design of \opt and the relative optimization mechanism of GRPO. 
For a given table image, each sampled recognition response is assigned a scalar reward defined as the average QA score over the associated QA set. 
GRPO does not optimize these rewards in absolute terms; instead, it updates the policy according to the relative advantage of each response within the sampled group. 
Consequently, what primarily matters is whether a QA pair changes the reward ordering among candidate responses, rather than its standalone score.

When a QA pair is incorrect or ill-posed, it is unlikely to be answered correctly by any of the sampled responses, regardless of the recognition quality. 
In this case, the corresponding QA contributes nearly the same low score to all responses in the group. 
Such a shared contribution acts approximately as a common offset in the aggregated reward and is therefore largely removed during the within-group normalization in GRPO. 
As a result, these noisy QA pairs have limited influence on the relative preference structure among candidate responses, and thus do not substantially alter the optimization direction.

From this perspective, the main effect of noisy QA pairs is not to introduce systematic bias, but to reduce the number of informative reward components available for estimating response quality.
Our proposed Attention-guided QA generation aims to mitigate this by providing more reliable reward estimation with diverse QA samples.

\subsection{Effect of Stage-2 Data Size}
\label{supp_sec:stage2_size}

We also investigate the relationship between Stage-2 supervised fine-tuning data size and the final performance of \opt. 
As shown in Figure~\ref{fig:x3HF-2}, strong performance can already be achieved with a relatively small amount of Stage-2 data.

This result suggests that the main role of Stage 2 is to provide a stable and reliable policy initialization for subsequent GRPO optimization, rather than to serve as the primary source of performance gains. 
Once a reasonable initialization is obtained, the Stage-3 \opt training contributes most of the improvement.

\subsection{Results on Standard Large-Scale Benchmarks}
\label{supp_sec:large_benchmarks}

\begin{table}[h!]\setlength\tabcolsep{2.5pt}\footnotesize
    	\centering
    	\vspace{-0.5em}
    	\begin{tabular}{|l|cc|cc|cc|}
    		\hline
    		\multirow{2}{*}{\textbf{\begin{tabular}[c]{@{}l@{}}TR\\ Method\end{tabular}}}                    & \multicolumn{2}{c|}{\textbf{Complex}*} & \multicolumn{2}{c|}{\textbf{PubTabNet}}              & \multicolumn{2}{c|}{\textbf{FinTabNet}}                    \\ \cline{2-7} 
    		& \multicolumn{1}{c|}{\textbf{TEDS}} & \textbf{S-TEDS} & \multicolumn{1}{c|}{\textbf{TEDS}} & \textbf{S-TEDS}      & \multicolumn{1}{c|}{\textbf{TEDS}} & \textbf{S-TEDS} \\ \hline
    		SLANNet-plus      & \multicolumn{1}{c|}{68.19} & 79.21 & \multicolumn{1}{c|}{86.57}         & {96.43}  & \multicolumn{1}{c|}{73.77} & 84.84                                    \\ \hline
    		UniTable             & \multicolumn{1}{c|}{70.86} & 80.81 & \multicolumn{1}{c|}{86.44}         & 95.66           & \multicolumn{1}{c|}{75.15} & 82.97                                 \\ \hline
    		MuTabNet          & \multicolumn{1}{c|}{59.24} & 77.41  & \multicolumn{1}{c|}{94.42}        & 97.51           & \multicolumn{1}{c|}{97.24}  & 98.17  \\ \hhline{|=|=|=|=|=|=|=|}
    		TRivia (Stage-1)    & \multicolumn{1}{c|}{77.85}         & 83.23  & \multicolumn{1}{c|}{95.52}         & 96.43           & \multicolumn{1}{c|}{96.74} & 97.68                 \\ \hline
    		TRivia (Stage-2)   & \multicolumn{1}{c|}{88.57}         & 92.48 & \multicolumn{1}{c|}{90.81}         & 93.04           & \multicolumn{1}{c|}{89.60} & 91.82                   \\ \hline
    		TRivia-3B     & \multicolumn{1}{c|}{89.88}         & 93.60 & \multicolumn{1}{c|}{{91.79}} & 93.81           & \multicolumn{1}{c|}{91.66} & 93.71                \\ \hline
    		TRivia (SFT)     & \multicolumn{1}{c|}{89.58}         & 93.24   & \multicolumn{1}{c|}{{95.00}} & 96.05           & \multicolumn{1}{c|}{96.52} & 97.63                   \\ \hline
    	\end{tabular}\\\vspace{0.4em}
    	{* 1,512 complex tables from OmniDocBench, CC-OCR, and OCRBench.}
    	\vspace{-0.5em}
    	\caption{TEDS on Complex and Standard Large Benchmarks.}\label{tab:KHxh-234}\vspace{-0.5em}
    \end{table}

To complement the main experiments on challenging real-world tables, we further evaluate \opt on the standard large-scale benchmarks PubTabNet and FinTabNet, and compare it with representative table recognition systems, including MuTabNet~\cite{ICDAR24KAT}.
The results are reported in Table~\ref{tab:KHxh-234}.

The comparison reveals two consistent observations. 
First, \opt-3B remains substantially stronger on complex real-world tables collected from OmniDocBench, CC-OCR, and OCRBench, where existing specialized models exhibit a clear performance gap. 
Second, because PubTabNet and FinTabNet are only used in Stage 1, the subsequent stages of real-world adaptation may reduce in-domain performance on these benchmarks. 
When these benchmark-style samples are incorporated again in the final supervised stage, the resulting model restores strong performance on PubTabNet and FinTabNet while preserving the advantage on complex tables. 
These results suggest that \opt is particularly beneficial for improving robustness to diverse real-world table layouts, while remaining compatible with conventional benchmark-oriented training when stronger in-domain accuracy is desired.

\onecolumn

\section{Prompt Template}
\label{supp_sec:prompt}
This section provides all prompt designs used in this paper.

Table~\ref{tab:qa_gen_prompt} shows the prompts for QA generation. 
We include several heuristic constraints in it.
For example, the questions should avoid direct references to rows or columns, such as "the third row", which is heavily limited to the structural parsing ability of the $M_{\text{QA}}$ model.
The questions should be simple, avoiding complex reasoning, to reduce the impact of $M_{\text{LLM}}$'s capability on reward estimation.

Since our task involves bilingual table recognition in Chinese and English, prompts for both question answering with the LLM and the VLM are prepared in both languages. 
This prevents the model from outputting answers in the wrong language, as shown in Table~\ref{tab:llm_qa_prompt}.

Table~\ref{tab:qa_val_prompt} shows the prompts that $M_{\text{Val}}$ use for cross-check filtering.

Finally, the table recognition prompt for the general-purpose VLM is shown in Table~\ref{tab:tab_recog_prompt}, which is developed through multiple iterations of optimization for best performance.

\begin{table*}[htbp]
    \centering
    \begin{tabular}{p{0.9\linewidth}}
    \toprule
    <image>\\
    Given an image, your task is to generate 10 reasonable and natural QA pairs based on the following rules: \\
    \\
    1. The questions should be contextually appropriate and natural. \\
    2. The answer to each question must be a short word or two or a numerical value. \\
    3. For tables in Chinese, generate QA pairs in Chinese. Ensure the question and answer are both in the same language. \\
    4. Each question should have one and only one answer. \\
    5. Distribute the QA pairs across different parts of the table to cover multiple data points, avoiding any concentration on a single row or column. \\
    6. Avoid questions that involve reasoning, such as numerical comparisons, maximum/minimum values, or calculations. \\
    7. Do not directly mention the table structure in the question; instead, incorporate natural references. For example, instead of asking, "What is the journal account in the first row?" ask, "What is the journal account for serial number 1?" \\
    8. Exclude questions that could be answered by only one data point, such as "Does the table include future goals?" or "Does the table list the budget?" \\
    9. Ensure the questions can be answered using an HTML-formatted table, and avoid referencing visual orientation or relative positioning (e.g., "What is to the left of a T-account?"). \\
    \\
    If you cannot generate QA pairs that meet the above criteria, output "None". Otherwise, output the generated QAs in JSON format. \\
    \\
    **Example Output Format:**\\
    ```json \\
    \quad [ \\
    \quad\quad {"question": "What is the market cap in Rmb mn?", "answer": "13,650.6"}, \\
    \quad\quad {"question": "What is the 12 month price target?", "answer": "24.80"}, \\
    \quad ] \\
    ``` \\
    \bottomrule
    \end{tabular}
    \caption{Q\&A Generation Prompt}
    \label{tab:qa_gen_prompt}
\end{table*}

\begin{table*}[htbp]
    \centering
    \begin{tabular}{p{0.9\linewidth}}
    \toprule
    \sethlcolor{yellow}\hl{\textbf{For question in English:}} \\
    Given an HTML-formatted table and a corresponding question, your task is to respond appropriately based on table. If the table do not contain the answer of question, output "Not answerable". \\
    Your answer should be a short phrase of only few words. Output the answer within <answer> </answer>. \\
    \\
    HTML Table: \textcolor{blue}{\{html\_table\}} \\
    \\
    Question: \textcolor{blue}{\{question\}} \\
    \midrule
    \sethlcolor{yellow}\hl{\textbf{For question in Chinese:}} \\
    \begin{CJK*}{UTF8}{gbsn}给定一个HTML格式的表格以及一个相应的问题，你的任务是根据表格回答该问题。如果该表格不包含该问题的答案，请输出"无法回答"。你的答案必须简短、仅有一两个词语。输出答案时用<answer></answer>包裹。\end{CJK*} \\
    \\
    \begin{CJK*}{UTF8}{gbsn}HTML表格: \textcolor{blue}{\{html\_table\}}\end{CJK*} \\
    \\
    \begin{CJK*}{UTF8}{gbsn}问题: \textcolor{blue}{\{html\_table\}}\end{CJK*} \\
    \bottomrule
    \end{tabular}
    \caption{LLM QA Prompt}
    \label{tab:llm_qa_prompt}
\end{table*}

\begin{table*}[htbp]
    \centering
    \begin{tabular}{p{0.9\linewidth}}
    \toprule
    \sethlcolor{yellow}\hl{\textbf{For question in English with image input:}} \\
    <image>\\
    Given a table image and a corresponding question, your task is to respond appropriately based on table image. If the table do not contain the answer of question, output "Not answerable". \\
    Your answer should be a short phrase of only few words. Output the answer within <answer> </answer>. \\
    \\
    Question: \textcolor{blue}{\{question\}} \\
    \midrule
    \sethlcolor{yellow}\hl{\textbf{For question in Chinese with image input:}} \\
    <image>\\
    \begin{CJK*}{UTF8}{gbsn}给定一个表格图像以及一个相应的问题，你的任务是根据表格图片回答该问题。如果该表格不包含该问题的答案，请输出"无法回答"。你的答案必须简短、仅有一两个词语。输出答案时用<answer></answer>包裹。\end{CJK*} \\
    \\
    \begin{CJK*}{UTF8}{gbsn}问题: \textcolor{blue}{\{question\}}\end{CJK*} \\
    \sethlcolor{yellow}\hl{\textbf{For question in English without image input:}} \\
    Answer the following question. Your answer should be a short phrase of only few words. If you cannot answer this question, output "Not answerable". \\
    Your answer should be a short phrase of only few words. Output the answer within <answer> </answer>. \\
    \\
    Question: \textcolor{blue}{\{question\}} \\
    \sethlcolor{yellow}\hl{\textbf{For question in Chinese without image input:}} \\
    \begin{CJK*}{UTF8}{gbsn}回答下面的问题。你的答案必须简短、仅有一两个词语。如果你无法回答该问题，请输出"无法回答"。你的答案必须简短、仅有一两个词语。输出答案时用<answer></answer>包裹。\end{CJK*} \\
    \\
    \begin{CJK*}{UTF8}{gbsn}问题: \textcolor{blue}{\{question\}}\end{CJK*} \\
    \bottomrule
    \end{tabular}
    \caption{VLM QA cross-checking Prompt}
    \label{tab:qa_val_prompt}
\end{table*}

\begin{table*}[htbp]
    \centering
    \begin{tabular}{p{0.9\linewidth}}
    \toprule
    \sethlcolor{yellow}\hl{\textbf{For general purpose VLMs:}} \\
    <image>\\
    You are an AI specialized in recognizing and extracting table from images. Your mission is to analyze the table image and generate the result in HTML format using specified tags. Output only the results without any other words and explanation. \\
    \sethlcolor{yellow}\hl{\textbf{For TRivia-3B:}} \\
    <image>\\
    You are an AI specialized in recognizing and extracting table from images. Your mission is to analyze the table image and generate the result in OTSL format using specified tags. Output only the results without any other words and explanation. \\
    \bottomrule
    \end{tabular}
    \caption{Table Recognition Prompt}
    \label{tab:tab_recog_prompt}
\end{table*}

\end{document}